\pdfoutput=1

\documentclass[11pt]{article}

\usepackage[]{twocol}

\usepackage{times}
\usepackage{latexsym}

\usepackage[T1]{fontenc}

\usepackage[utf8]{inputenc}

\usepackage{microtype}

\usepackage[utf8]{inputenc}
\usepackage{amsmath}
\usepackage{graphicx}
\usepackage{amsfonts}
\usepackage{amssymb}
\usepackage{booktabs}
\usepackage{multirow}
\usepackage{url}
\usepackage[nameinlink,capitalize]{cleveref}

\usepackage{authblk}

\usepackage{acro}
\DeclareAcronym{art}{short = ART, long  = approximate randomization testing, cite = Riezler05}
\DeclareAcronym{bleu}{short = BLEU, long  = bilingual evaluation understudy, cite = Papineni02}
\DeclareAcronym{bpe}{short = BPE, long  = byte pair encoding, cite = Sennrich16a}
\DeclareAcronym{cat}{short = CAT, long  = computer-aided translation}
\DeclareAcronym{cbmt}{short = CBMT, long  = character-based machine translation}
\DeclareAcronym{cbsmt}{short = CBSMT, long  = character-based statistical machine translation}
\DeclareAcronym{cbnmt}{short = CBNMT, long  = character-based neural machine translation}
\DeclareAcronym{cm}{short = CM, long  = confidence measures}
\DeclareAcronym{cer}{short = CER, long  = character error rate}
\DeclareAcronym{fda}{short = FDA, long  = feature decay algorithm, cite = Biccici15}
\DeclareAcronym{hmm}{short = HMM, long  = hidden Markov alignment models, cite = Vogel96}
\DeclareAcronym{htr}{short = HTR, long  = handwritten text recognition}
\DeclareAcronym{il}{short = IL, long  = incremental learning}
\DeclareAcronym{imt}{short = IMT, long  = interactive machine translation}
\DeclareAcronym{inmt}{short = INMT, long  = interactive neural machine translation}
\DeclareAcronym{ismt}{short = ISMT, long  = interactive statistical machine translation}
\DeclareAcronym{ksr}{short = KSR, long  = key stroke rate}
\DeclareAcronym{lstm}{short = LSTM, long  = long short-term memory, cite = Hochreiter97}
\DeclareAcronym{mar}{short = MAR, long  = mouse action rate}
\DeclareAcronym{mt}{short = MT, long  = machine translation}
\DeclareAcronym{nlp}{short = NLP, long  = natural language processing}
\DeclareAcronym{nmt}{short = NMT, long  = neural machine translation}
\DeclareAcronym{pe}{short = PE, long  = post-editing}
\DeclareAcronym{rbmt}{short = RBMT, long  = rule-based machine translation}
\DeclareAcronym{relu}{short = ReLU, long  = rectified linear unit}
\DeclareAcronym{rnn}{short = RNN, long  = recurrent neural network, cite = {Hochreiter97}}
\DeclareAcronym{sgd}{short = SGD, long  = stochastic gradient descend}
\DeclareAcronym{smt}{short = SMT, long  = statistical machine translation}
\DeclareAcronym{ter}{short = TER, long  = translation error rate, cite = Snover06}
\DeclareAcronym{wer}{short = WER, long  = word error rate}
\DeclareAcronym{wlac}{short = WLAC, long  = word level autocompletion, cite = Li21}
\DeclareAcronym{wsr}{short = WSR, long  = word stroke rate}
\DeclareAcronym{xml}{short = XML, long  = eXtensible Markup Language}

\makeatletter
\newcommand\footnoteref[1]{\protected@xdef\@thefnmark{\ref{#1}}\@footnotemark}
\makeatother

\title{Findings of the Covid-19 MLIA Machine Translation Task}
\begin{document}
	\author[1]{\vspace{-2pt} Francisco Casacuberta}
	\author[2]{Alexandru Ceausu}
	\author[3]{Khalid Choukri}
	\author[4]{Miltos Deligiannis}
	\author[1]{Miguel Domingo}
	\author[5]{\protect\\ \vspace{-2pt}Mercedes Garc{\'i}a-Mart{\'i}nez}
	\author[5]{Manuel Herranz}
	\author[6]{Guillaume Jacquet}
	\author[4]{Vassilis Papavassiliou}
	\author[4]{\protect\\ Stelios Piperidis}
	\author[4]{Prokopis Prokopidis}
	\author[4]{Dimitris Roussis}
	\author[3]{Marwa Hadj Salah}
	\affil[1]{Universitat Polit{\`e}cnica de Val{\`e}ncia \texttt{\small\{fcn,midobal\}@prhlt.upv.es}}
	\affil[2]{Court of Justice of the EU \texttt{\small Alexandru.Ceausu@curia.europa.eu}}
	\affil[3]{Evaluations and Language resources Distribution Agency \texttt{\small\{choukri,marwa\}@elda.org}}
	\affil[4]{Athena Research Center \texttt{\small\{mdel,vpapa,spip,prokopis\}@athenarc.gr}}
	\affil[5]{Pangeanic S. L. - PangeaMT \texttt{\small\{m.garcia,m.herranz\}@pangeanic.com}}
	\affil[6]{European Commission \texttt{\small guillaume.jacquet@ec.europa.eu}}
\maketitle
\begin{abstract}
This work presents the results of the machine translation (MT) task from the Covid-19 MLIA @ Eval initiative, a community effort to improve the generation of MT systems focused on the current Covid-19 crisis. Nine teams took part in this event, which was divided in two rounds and involved seven different language pairs. Two different scenarios were considered: one in which only the provided data was allowed, and a second one in which the use of external resources was allowed. Overall, best approaches were based on multilingual models and transfer learning, with an emphasis on the importance of applying a cleaning process to the training data.
\end{abstract}

\section{Introduction}
\label{se:intro}
In the current Covid-19 crisis, as in many other emergency situations, the general public, as well as many other stakeholders, need to aggregate and summarize different sources of information into a single coherent synopsis or narrative, complementing different pieces of information, resolving possible inconsistencies and preventing misinformation. This should happen across multiple languages, sources and levels of linguistic knowledge that varies depending on social, cultural or educational factors.

The Covid-19 MLIA @ Eval initiative\footnote{\url{http://eval.covid19-mlia.eu/}} consisted of three Natural Language Processing tasks: (1) information extraction, (2) multilingual semantic search and (3) machine translation. 
The goal was to organize a community evaluation effort aimed at accelerating the creation of resources and tools for improving the deployment of automatic systems focused on Covid-19 related documents.

In this paper, we focus on the 3rd task about \ac{mt}. 
The task was divided in two rounds. At the end of each round, participants wrote or updated their report describing their system and highlighting which methods and data had been used.

The main event for machine translation is the Workshop/conference of Machine Translation (WMT\footnote{\url{https://www.statmt.org/}}) where several  machine translation competition are organized since 2006. A competition\footnote{\url{https://www.statmt.org/wmt14/medical-task/}.} on medical data was organized on 2014, and a biomedical one\footnote{\url{https://www.statmt.org/wmt22/biomedical-translation-task.html}.} is being organized since 2016. The conference is held annually in connection with larger conferences on natural language processing. 
Another important machine translation competition is the International Workshop of Spoken Language Translation\footnote{\url{https://iwslt.org}} (IWSLT) about spoken translation.  
The Covid-19 MLIA @ Eval \ac{mt} task focused on text from the Covid-19 crisis that shocked the world and for which there are not a lot of processed text or corpora.

\section{Task description}
\label{se:task}
The goal of the MLIA \ac{mt} task was to benchmark \ac{mt} systems focused on Covid-19 related documents for different language pairs. \cref{fi:examples} shows some examples of sentences from Covid-19 related documents.

\begin{figure}
	\emph{30\% of children and adults infected with measles can develop complications.} \\
	
	\emph{The MMR vaccine is safe and effective and has very few side effects.} \\
	
	\emph{The first dose is given between 10 and 18 months of age in European countries.} \\
	
	\emph{Note: The information contained in this factsheet is intended for the purpose of general information and should not be used as a substitute for the individual expertise and judgement of a healthcare professional.} \\
	\caption{Examples of English sentences from Covid-19 related documents.}
	\label{fi:examples}
\end{figure}

Given a set of training data provided by the organizers for each language pair, participants had to train up to five different \ac{mt} systems per language pair (see \cref{se:lp}). These systems were classified in two scenarios:

\begin{itemize}
	\item \textbf{Constrained}: systems could be trained exclusively with data provided by the organizers (including data from a different language pair, monolingual data, etc). The use of basic linguistic tools such as taggers, parsers or morphological analyzers or multilingual systems was allowed for this scenario.
	\item \textbf{Unconstrained}: systems could be trained using data not provided by the organizers, from any external resource not allowed in the constrained scenario.
\end{itemize}

Systems were evaluated and compared according to the scenario to which they belonged. It was mandatory that one of the submitted systems per language pair belonged to the constrained scenario. Participants were able to take part in any or all of the language pairs. They used their systems to translate a test set of unseen sentences in the source language. Evaluation consisted on assessing the translation quality of the submissions. Different metrics were used on each round.

\subsection{Language pairs}
\label{se:lp}
The following language pairs were addressed through out the initiative:
\begin{itemize}
	\item English{\textendash}German.
	\item English{\textendash}French.
	\item English{\textendash}Spanish.
	\item English{\textendash}Italian.
	\item English{\textendash}Modern Greek.
	\item English{\textendash}Swedish.
	\item English{\textendash}Arabic\footnote{This language pair was only used on the second round.}.
\end{itemize}
In all cases, the only translation direction was from English to the other language.

\subsection{Data generation}
\label{se:dg}
In the context of the first round of this initiative, we decided to collect an initial collection of parallel corpora in health and medicine domains from well-known web sources and enrich them with identified Covid-19 parallel data. The purpose of following this approach was to implement a very quick response of the \ac{mt} community in an emergency situation, like the current pandemic.

To this end, we first collected an updated version of the European Medicines Agency (EMEA) corpus\footnote{\url{https://www.prhlt.upv.es/~mt/prokopidis-and-papavassiliou-emea.html}.}, and applied new (more robust and efficient) methods for text extraction from \emph{pdf} files, sentence splitting, sentence alignment and parallel corpus filtering. Moreover, medical-related multilingual collections which were offered by the Publications Office of EU\footnote{\url{https://op.europa.eu/en/home}.} were processed in a similar manner, increasing the volume of the ``general'' subset of the training data.

The first step of acquiring Covid-19-related data was the identification of several bilingual websites with such content. With the aim of constructing datasets that could be publicly available, we targeted websites of national authorities and public health agencies\footnote{Such a list is available at \url{https://www.ecdc.europa.eu/en/COVID-19/national-sources}.}, EU agencies and specific broadcast websites (e.g., voxeurop\footnote{\url{https://voxeurop.eu/}.}, GlobalVoices\footnote{\url{https://globalvoices.org/}.} or Voltairnet\footnote{\url{https://www.voltairenet.org/}.}).

For acquiring domain-specific bilingual corpora, we used a recent version of ILSP-FC \citep{Papavassiliou13}, a modular toolkit that integrates modules for text normalization, language identification, document clean-up, text classification, bilingual document alignment (i.e., identification of pairs of documents that are translations of each other) and sentence alignment. As mentioned above, taking into account the emergency situation, a ``rapid'' approach based on keywords was adopted for text classification (i.e., keeping only documents that are strongly related to the current worldwide health crisis). Specifically for sentence alignment, the LASER\footnote{\url{https://github.com/facebookresearch/LASER}.} toolkit was used instead of the integrated aligner. Then, a battery of criteria was applied on aligned sentences to automatically filter out sentence pairs with potential alignment or translation issues (e.g., with score less than a predefined threshold) or of limited use for training \ac{mt} systems (e.g., duplicate pairs, identical segments in a pair, etc.) and, thus, generate precision-high language resources.

For the second round, we repeated the previous process{\textemdash}re-crawling several websites of national authorities and public health agencies{\textemdash}in order to enrich the data that had already been collected. Additionally, we exploited the outcomes of an available infrastructure, namely MediSys\footnote{\url{https://jeodpp.jrc.ec.europa.eu/ftp/jrc-opendata/LANGUAGE-TECHNOLOGY/EMM_collection/2020_MediSys_Covid19_dataset/}.} (Medical Information System), with the purpose of constructing parallel corpora beneficial for \ac{mt} \citep{Roussis22}. Similarly to the first round's approach, it could be seen as an application of implementing a quick response of the \ac{mt} community to the pandemic crisis. 

MediSys is one of the publicly accessible systems of the Europe Media Monitor (EMM) which processes media to identify potential public health threats in a fully automated fashion \citep{Linge10}. Focusing on the current pandemic, a dataset of metadata which concerns Covid-19 related news was made publicly available in RSS/XML format, which corresponded to millions of news articles \citep{Jacquet20}. The dataset was divided into subsets according to the articles' month of publication. First, the metadata were parsed and the URL and language of each article were extracted. Then, each web page of the targeted languages was fetched and its main content was stored in a text file. The generated text files were merged to create a single document for each language and each period. Thus, these documents constituted the Covid-19 related monolingual corpora and were considered comparable (in pairs), due to their narrow topic and the fact that they were published in the same time period. To this end, the LASER toolkit was applied on each document pair to mine sentence alignments for each EN-X language pair. Finally, several filtering methods were adopted (i.e., thresholding the alignment score by 1.04, removing near de-duplicates, etc) to compile the final dataset.

\subsection{Corpora}
\label{se:corp}
For the first round, we selected the data described in the previous section (\cref{se:dg}) and split them into train, validation and test.
Then, to ensure that the tests were a good representation of the task and were appropriate for being used for evaluation, we sorted all segments from the initial test according to the alignment probability between source and target. After that, we filtered them according to their number of words: removing those segments whose source had either less than $0.7$ or more than $1.3$ times the average number of words per sentence from the training set. Finally, we selected the first two thousand segments to construct the final version of the test set for round 1.

This process was improved for selecting round 2's corpora. Given the data used for this round, we computed some statistics and removed the outliers (segments that contained more than 100 words in either its source or target). Then, we split the data into train, validation and test sets. Since the data came from different sources, we wanted to ensure that both the validation and tests sets were representative enough of the training sets. For this reason, for each language pair, we computed the representation of each source in the total data (i.e., the number of segments from this source divided by the total number of segments). Then, out of the total segments we wanted to select for validation and test (4000 for each), we select that same percentage from each source.

Additionally, to ensure that validation and test did not contain low-quality segments (given that the data had been crawled from the web), we sorted the segments according to its alignment quality. Finally, we shuffled the selected segments and split them equally into validation and test.

Therefore, the procedure we followed for each language pair was:

\begin{enumerate}
	\item We computed the ratio of data from each different source over the total data.
	\item We computed the average number of words per segment over this set.
	\item We constituted a subset [0.7 * average words per segment, 1.3 * average words per segment].
	\item We sorted this subset (from best to worst) according to its alignment score.
	\item We selected the best 8000 * the percentage obtained at step 1 segments.
	\item We shuffled those segments and select half of them for validation and the other half for test.
\end{enumerate}

\cref{ta:corp} describes the corpora statistics.

\begin{table*}[!ht]
	\caption{Corpora statistics, divided by rounds. $|S|$ stands for number of sentences, $|T|$ for number of tokens and $|V|$ for size of the vocabulary. M denotes millions and K thousands.}
	\label{ta:corp}
	\resizebox{0.72\textwidth}{!}{\begin{minipage}{\textwidth}
			\centering
			\begin{tabular}{c c c c c c c c c c c c c c}
				\multicolumn{14}{c}{\textbf{Round 1}} \\
				\toprule
				&  & \multicolumn{2}{c}{\textbf{German}} & \multicolumn{2}{c}{\textbf{French}} & \multicolumn{2}{c}{\textbf{Spanish}} & \multicolumn{2}{c}{\textbf{Italian}} & \multicolumn{2}{c}{\textbf{Modern Greek}} & \multicolumn{2}{c}{\textbf{Swedish}} \\
				\cmidrule(lr){3-4}\cmidrule(lr){5-6}\cmidrule(lr){7-8}\cmidrule(lr){9-10}\cmidrule(lr){11-12}\cmidrule(lr){13-14}
				&  & \textbf{En} & \textbf{De} & \textbf{En} & \textbf{Fr} & \textbf{En} & \textbf{Es} & \textbf{En} & \textbf{It} & \textbf{En} & \textbf{El} & \textbf{En} & \textbf{Sv} \\
				\midrule
				\multirow{3}{*}{Train} & $|S|$ & \multicolumn{2}{c}{926.6K} & \multicolumn{2}{c}{1.0M} & \multicolumn{2}{c}{1.0M} & \multicolumn{2}{c}{900.9K} & \multicolumn{2}{c}{834.2K} & \multicolumn{2}{c}{806.9K} \\
				& $|T|$ & 17.3M & 16.1M & 19.4M & 22.6M & 19.5M & 22.3M & 16.7M & 18.2M & 15.0M & 16.4M & 14.5M & 13.2M \\
				& $|V|$ & 372.2K & 581.6K & 401.0K & 438.9K & 404.4K & 458.0K & 347.7K & 416.0K & 305.7K & 407.5K & 298.2K & 452.0K \\
				\midrule
				\multirow{3}{*}{Validation} & $|S|$ & \multicolumn{2}{c}{528} & \multicolumn{2}{c}{728} & \multicolumn{2}{c}{2.5K} & \multicolumn{2}{c}{3.7K} & \multicolumn{2}{c}{3.9K} & \multicolumn{2}{c}{723} \\
				& $|T|$ & 8.2K & 7.6K & 17.0K & 18.8K & 48.9K & 56.2K & 78.2K & 84.0K & 73.0K & 72.7K & 11.4K & 10.0K \\
				& $|V|$ & 2.4K & 2.6K & 4.1K & 4.5K & 9.7K & 10.6K & 12.4K & 14.9K & 10.3K & 14.5K & 2.6K & 2.8K \\
				\midrule
				\multirow{3}{*}{Test} & $|S|$ & \multicolumn{2}{c}{2000} & \multicolumn{2}{c}{2000} & \multicolumn{2}{c}{2000} & \multicolumn{2}{c}{2000} & \multicolumn{2}{c}{2000} & \multicolumn{2}{c}{2000} \\
				& $|T|$ & 34.9K & 33.2K & 33.2K & 35.8K & 32.6K & 34.3K & 33.7K & 34.2K & 42.6K & 44.3K & 35.3K & 30.6K \\
				& $|V|$ & 7.8K & 9.6K & 6.7K & 7.7K & 6.7K & 7.9K & 8.6K & 10.4K & 9.5K & 12.5K & 7.1K & 8.2K \\
				\bottomrule
			\end{tabular}
	\end{minipage}}
	
	\vspace{5pt}
	
	\resizebox{0.63\textwidth}{!}{\begin{minipage}{\textwidth}
			\centering
			\begin{tabular}{c c c c c c c c c c c c c c c c}
				\multicolumn{16}{c}{\textbf{Round 2}} \\
				\toprule
				&  & \multicolumn{2}{c}{\textbf{German}} & \multicolumn{2}{c}{\textbf{French}} & \multicolumn{2}{c}{\textbf{Spanish}} & \multicolumn{2}{c}{\textbf{Italian}} & \multicolumn{2}{c}{\textbf{Modern Greek}} & \multicolumn{2}{c}{\textbf{Swedish}} & \multicolumn{2}{c}{\textbf{Arabic}} \\
				\cmidrule(lr){3-4}\cmidrule(lr){5-6}\cmidrule(lr){7-8}\cmidrule(lr){9-10}\cmidrule(lr){11-12}\cmidrule(lr){13-14}\cmidrule(lr){15-16}
				&  & \textbf{En} & \textbf{De} & \textbf{En} & \textbf{Fr} & \textbf{En} & \textbf{Es} & \textbf{En} & \textbf{It} & \textbf{En} & \textbf{El} & \textbf{En} & \textbf{Sv} & \textbf{En} & \textbf{Ar} \\
				\midrule
				\multirow{3}{*}{Train} & $|S|$ & \multicolumn{2}{c}{1.5M} & \multicolumn{2}{c}{2.4M} & \multicolumn{2}{c}{2.9M} & \multicolumn{2}{c}{1.0M} & \multicolumn{2}{c}{674.0K} & \multicolumn{2}{c}{375.0K} & \multicolumn{2}{c}{424.4K} \\
				& $|T|$ & 23.5M & 22.1M & 45.6M & 53.0M & 52.4M & 60.3M & 16.4M & 17.2M & 11.4M & 12.2M & 5.5M & 5.1M & 7.7M & 7.5M \\
				& $|V|$ & 523.9K & 847.5K & 782.2K & 781.4K & 850.0K & 950.2K & 421.2K & 501.3K & 289.7K & 378.7K& 180.7K & 234.7K & 222.2K & 360.2K \\
				\midrule
				\multirow{3}{*}{Validation} & $|S|$ & \multicolumn{2}{c}{4.0K} & \multicolumn{2}{c}{4.0K} & \multicolumn{2}{c}{4.0K} & \multicolumn{2}{c}{4.0K} & \multicolumn{2}{c}{4.0K} & \multicolumn{2}{c}{4.0K} & \multicolumn{2}{c}{4.0K} \\
				& $|T|$ & 62.2K & 61.2K & 72.0K & 83.9K & 72.2K & 81.4K & 64.6K & 69.0K & 67.8K & 72.5K & 56.6K & 54.4K & 75.9K & 74.7K \\
				& $|V|$ & 13.9K & 17.1K & 13.2K & 14.8K & 13.8K & 15.8K & 14.6K & 16.7K & 14.0K & 18.0K & 12.3K & 14.1K & 16.1K & 23.7K \\
				\midrule
				\multirow{3}{*}{Test} & $|S|$ & \multicolumn{2}{c}{4.0K} & \multicolumn{2}{c}{4.0K} & \multicolumn{2}{c}{4.0K} & \multicolumn{2}{c}{4.0K} & \multicolumn{2}{c}{4.0K} & \multicolumn{2}{c}{4.0K} & \multicolumn{2}{c}{4.0K} \\
				& $|T|$ & 62.2K & 61.0K & 72.3K & 84.1K & 72.2K & 81.4K & 64.3K & 68.7K & 67.8K & 72.4K & 56.5K & 54.3K & 76.1K & 74.5K \\
				& $|V|$ & 13.8K & 17.0K & 13.1K & 14.8K & 13.7K & 15.7K & 14.4K & 16.7K & 14.1K & 18.2K & 12.3K & 14.1K & 16.2K & 23.5K \\
				\bottomrule
			\end{tabular}
	\end{minipage}}
\end{table*}

\subsection{Evaluation}
\label{se:eval1}
In order to evaluate the participant's systems, we selected the \ac{bleu}{\textemdash}which computes the geometric average of the modified n-gram precision, multiplied by a brevity factor{\textemdash}as our main metric, using \texttt{sacreBLEU} \citep{Post18} to compute it ensuring consistent scores. 

Additionally, we selected different alternative well-known \ac{mt} metrics for each round:

\begin{itemize}
	\item Round 1:
	\begin{description}
		\item[Character n-gram F-score (ChrF)] \citep{Popovi15}: character n-gram precision and recall arithmetically averaged over all character n-grams.
	\end{description}
	
	\item Round 2:
	\begin{description}
		\item[Translation Edit Rate (TER)] \citep{Snover06}: this metric computes the number of word edit operations (insertion, substitution, deletion and swapping), normalized by the number of words in the final translation.
		\item[BEtter Evaluation as Ranking (BEER)] \citep{Stanojevic14}: a sentence level metric that incorporates a large number of features combined in a linear model.
	\end{description}
\end{itemize}

We applied \ac{art}{\textemdash}with $10,000$ repetitions and using a $p$-value of $0.05${\textemdash}to determine whether two systems presented statistically significance. The scripts used for conducting the automatic evaluation are publicly available together with some utilities which were useful for the shared task\footnote{\url{https://github.com/midobal/covid19mlia-mt-task}.}.

Following the WMT criteria \citep{Barrault20}, we grouped systems together into clusters according to the statistical significance of their performance (as determined by \ac{art}). With that purpose, we sorted the submissions according to each metric and computed the significance of the performance between one system and the following. If it was not significant, we added the second system into the cluster of the first system\footnote{Considering that, at the start of this process, there is an initial cluster containing the first system.}. Otherwise, we added it into a new cluster. This way, systems from one cluster significantly outperformed all others in lower ranking clusters.

\subsection{Baselines}
\label{se:base}
At each round, we trained two different baselines to have an estimation of the expected translation quality of each scenario. 

\subsection*{Round 1}
On our first approach to the initiative, we trained two different constrained systems to use as baselines: one based on recurrent neural networks (RNN) \citep{Sutskever14,Bahdanau15} and another one based on the Transformer architecture \citep{Vaswani17}. All systems were built using \texttt{OpenNMT-py} \citep{Klein17}.

Systems for the RNN baselines were trained using the standard parameters: long short-term memory units \citep{Gers00}, with all model dimensions set to $512$; Adam \citep{Kingma14}, with a fixed learning rate of $0.0002$ and a batch size of $60$; label smoothing of $0.1$ \citep{Szegedy15}; beam search with a beam size of $6$; and joint \ac{bpe} applied to all corpora, using $32,000$ merge operations.

Similarly, systems for the Transformer baselines were trained using the standard parameters: $6$ layers; Transformer \citep{Vaswani17}, with all dimensions set to $512$ except for the hidden transformer feed-forward (which was set to $2048$); $8$ heads of Transformer self-attention; $2$ batches of words in a sequence to run the generator on in parallel; a dropout of $0.1$; Adam \citep{Kingma14}, using an Adam beta2 of $0.998$, a learning rate of $2$ and Noam learning rate decay with $8000$ warm up steps; label smoothing of $0.1$ \citep{Szegedy15}; beam search with a beam size of $6$; and joint \ac{bpe} applied to all corpora, using $32,000$ merge operations.

\subsection*{Round 2}
Given the results from the previous round, we decided to drop the RNN architecture and use only the Transformer one. Thus, we trained a system using only the second round data and another using also the first round data (except for En{\textendash}Ar, for which only round 2 data was available).

Like in round 1, systems were also built using \texttt{OpenNMT-py} and were based on the Transformer architecture. They were trained using the standard parameters: $6$ layers; Transformer \citep{Vaswani17}, with all dimensions set to $512$ except for the hidden transformer feed-forward (which was set to $2048$); $8$ heads of Transformer self-attention; $2$ batches of words in a sequence to run the generator on in parallel; a dropout of $0.1$; Adam \citep{Kingma14}, using an Adam beta2 of $0.998$, a learning rate of $2$ and Noam learning rate decay with $8000$ warm up steps; label smoothing of $0.1$ \citep{Szegedy15}; beam search with a beam size of $6$; and joint \ac{bpe} applied to all corpora, using $32,000$ merge operations.

\section{Participants}
In this section, we present the list of participants that took part in this event, ordered alphabetically according to their affiliation:

\begin{itemize}
	\item \textbf{Accenture}\footnote{\url{https://www.accenture.com}.}: Paul Rodrigues.
	\item \textbf{Translation Centre for the Bodies of the European Union (CdT-ASL)}\footnote{\url{https://cdt.europa.eu/en}.}: Zuzanna Parcheta, Daniel Mar{\'i}n Buj and Anna Samiotou.
	\item \textbf{Charles University - Machine Translation (CUNI-MT)}\footnote{\label{CUNI}\url{https://ufal.mff.cuni.cz}.}: Ivana Kvapil{\'i}ková and Ondřej Bojar.
	\item \textbf{Charles University - Machine Translation Information Retrieval (CUNI-MTIR)}\footnoteref{CUNI}: Shadi Saleh, Hashem Sellat, Hadi Abdi Khojasteh and Pavel Pecina.
	\item \textbf{E-Translation}\footnote{\url{https://ec.europa.eu/info/index_en}.}: Csaba Oravecz, Katina Bontcheva, David Kolovratn{\'i}k, Bogomil Kovachev, Vilmantas Liubinas, Christopher Scott, Francois Thunus and Andreas Eisele.
	\item \textbf{Lingua Custodia (LC)}\footnote{\url{https://www.linguacustodia.finance/en/welcome}.}: Raheel Qader.
	\item \textbf{Laboratoire d'Informatique pour la Mécanique et les Sciences de l'Ingénieur (LIMSI)}\footnote{\url{https://www.limsi.fr/fr}.}: Sadaf Abdul Rauf and François Yvon.
	\item \textbf{PROMT}\footnote{\url{https://www.promt.ru}.}: Alexander Molchanov.
	\item \textbf{TARJAMA-AI}\footnote{\url{https://www.tarjama.com}.}: Abdallah Nasir, Ruba Jaikat, Sara Alisis and Tamer Alnasser.
\end{itemize}

\subsection{Participants' approaches}
\label{se:participants}
In this subsection, we present the different approaches submitted by each team.

\subsubsection*{Accenture}
This team did not submitted a full report describing their approach. However, they noted in their system's description that they had used multilingual BART \citep{mbart}. They only participated in round 1.

\subsubsection*{CdT-ASL}
CdT-ASL team developed NICE which integrates neural machine translation custom engines for confidential adapted translations. They submitted constrained and unconstrained systems, they added generic and public health domains CdT own data for unconstrained systems. They applied cleaning processes to prepare the data for training with big transformer using \emph{OpenNMT-tf}. They only took part in round 2.

\subsubsection*{CUNI-MT}
For round 1, this team submitted 3 different approaches to the constrained scenario: standard \ac{nmt} with online back-translation \citep{onlineBT}; a transfer learning approach based on \citet{cunimttransfer}, who fine-tuned a low-resource child model from a pre-trained high-resource parent model for a different language pair; and a multilingual model{\textemdash}in which, during inference, the corresponding embedding of the target language was selected{\textemdash}trained to translate from English to French, Italian and Spanish (due to language similarities). The training was performed using the XLM\footnote{\url{https://github.com/facebookresearch/XLM}.} toolkit and the vocabulary size was set to 30k.

For round 2, they trained their multilingual models using Transformer in MarianNMT \citep{mariannmt}. They trained jointly on all languages. The results for the constrained scenario showed better results for transfer learning models. They concluded that when pre-training a model on a different language pair, better results are obtained when the corpus size is big and the transfer works also for completely unrelated languages.

\subsubsection*{CUNI-MTIR}
This team submitted systems for English into French, German, Swedish and Spanish in both constrained and unconstrained scenarios for round 1. The Transformer architecture from MarianNMT \citep{mariannmt} was used in order to train the models. For unconstrained systems, they used UFAL Medical Corpus\footnote{\url{http://ufal.mff.cuni.cz/ufal\_medical\_corpus}.} for training data and then fine-tuned the models with the constrained data. All the data was tokenized using Khresmoi\footnote{\url{http://www.khresmoi.eu/assets/Deliverables/WP4/KhresmoiD412.pdf}.}'s tokenizer and, then, encoded using \ac{bpe} with 32K merges.

\subsubsection*{E-Translation}
For round 1, this teams submitted systems to several constrained and unconstrained language pairs. They used transfer learning and 12K size vocabulary created using SentencePiece over Transformer models trained with MarianNMT \citep{mariannmt}. For the unconstrained scenario, they submitted their WMT system and a new version of that system, fine-tuned with the constrained data.

For round 2, they participated in both constrained and unconstrained scenarios for 6 language pairs (all but Arabic). They performed a general clean-up including a language identifier and checking the match of the number of tokens in source and target to filter noisy segments. For Greek and Spanish they did not do pre- or post-processing, only sanity checking. They experimented with standard Transformer and big Transformer in MarianNMT \citep{mariannmt}. For the unconstrained scenario, they made use of the TAUS Corona Crisis Corpora, the OPUS EMEA Corpus and a health related subset of the Euramis dataset.

\subsubsection*{Lingua Custodia}
Lingua Custodia's submissions for round 1 consisted of a multilingual model able to translate from English to French, German, Spanish, Italian and Swedish; and individual translation models for English{\textendash}German and English{\textendash}French. They applied unigram SentencePiece for subword segmentation using a source and target shared vocabulary of 50K for individual models and 70K for multilingual models. Additionally, authors split the numbers character by character. For multilingual models, a language token is added to the source in order to indicate the target language. The English{\textendash}German multilingual model achieved much higher score than the English{\textendash}German single model. This improvement is not shown in the English{\textendash}French model.

For round 2, they participated in the constrained scenario. The pre-processing used was based on Moses' tokenizer and cleaning techniques such as removing much longer sentences comparing source and target lengths, replacement of consecutive spaces by one space. They used inline casing consisting of adding a tag with the casing information. They finally append the language token to each source sentence in the pre-processing in order to indicate the target language for multilingual models. Standard transformer architecture in Sockeye toolkit was used for training in multiple GPUs instead of Seq2SeqPy used previously because the data loading is more efficient and has better support for multiple GPUs. 

\subsubsection*{LIMSI}
This team submitted systems to English{\textendash}French constrained and unconstrained scenarios for round 1. \ac{bpe} with 32K vocabulary units was applied to the constrained system. They submitted four unconstrained systems: 1) one system build using an external in-domain biomedical corpora; 2) a system first trained on WMT14\footnote{\url{http://www.statmt.org/wmt14/translation-task.html}.} general data and fine-tuned on the shared task's corpus; 3) same as 2) but adding BERT \citep{zhu2020incorporating}; and 4) a system only trained with constrained data but computing the \ac{bpe} codes using all the external in-domain corpus. Their systems were trained using Transformer architecture from \textit{fairseq}\footnote{\url{https://fairseq.readthedocs.io/en/latest/models.html}.} (Facebook's seq-2-seq library).

\subsubsection*{PROMT}
For round 1, PROMT's approaches consisted in a multilingual model trained using MarianNMT's \citep{mariannmt} Transformer architecture. For the constrained scenario, all data was concatenated using de-duplication to one single multilingual corpus to build a 8k SentencePiece \citep{sentencepiece} model for subword segmentation. In addition, a language-specific tag was added to the source side of the parallel sentence pairs (e.g., \emph{$<it>$} token was added to the beginning of the English sentence of the English{\textendash}Italian sentence pair). They also removed all tokens that appeared less than ten times in the combined de-duplicated monolingual corpus from their vocabulary.

For the unconstrained scenario, all available data mainly from the OPUS \citep{opus} and statmt\footnote{\url{http://www.statmt.org/}.} with the addition of private data harvested from the Internet were added to the training data. A special \ac{bpe} implementation \citep{molchanov-2019-promt} developed by the team was applied instead of SentencePiece, but the authors used SentencePiece in the constrained scenario as it seemed to work better in low-resource settings. The size of the \ac{bpe} models and vocabularies varied from 8k to 16k and shared vocabulary was not used (separate \ac{bpe} models were trained) for the English{\textendash}Greek pair as the two languages have different alphabets.

For round 2, they trained a transformer multilingual model with a single encoder and a single decoder with Marian toolkit. For round 2, PROMT performed fine-tuning for the language pairs improving 1-2 additional BLEU points in constrained mode. For the unconstrained scenario, they used the same approach as in round 1.

\subsubsection*{TARJAMA-AI}
This team submitted a single system for English to Spanish, German, Italian, French and Swedish constrained scenario. This system consisted of a model trained with all the language pairs data adding a special token for the non-target languages. Additionally, they over-sampled the corpus of the desired target language (i.e., the English{\textendash}Spanish corpus for training the constrained English{\textendash}Spanish, etc). They only took part in round 1.

\section{Results}
\label{se:res}
In this section, we present the results from each round. Following the WMT criteria \citep{Barrault20}, we grouped systems
together into clusters according to which systems significantly outperformed all others in lower ranking clusters, according to \ac{art} (see \cref{se:eval1}). For clarity purposes, we use BLEU as the main metric for performing the ranking. Nonetheless, we tried using each metric from \cref{se:eval1} as the main one for ranking, observing that all of them presented a similar behavior.

\begin{table*}[!ht]
	\caption{Results of the first round, divided by categories. Systems are ranked according to BLEU. Lines indicate clusters according to ART. Systems within a cluster are considered tied and, thus, are ranked equally.}
	\label{ta:res1}
	\hspace{-20pt}
	\resizebox{0.7\textwidth}{!}{\begin{minipage}{\textwidth}
			\begin{tabular}{c c c c c c}
				\multicolumn{6}{c}{\textbf{English{\textendash}German}} \\
				\toprule
				& \textbf{Rank} & \textbf{Team} & \textbf{Description} & \textbf{BLEU [$\uparrow$]} & \textbf{chrF [$\uparrow$]} \\
				\cmidrule(lr){2-2}\cmidrule(lr){3-3}\cmidrule(lr){4-4}\cmidrule(lr){5-5}\cmidrule(lr){6-6}
				\parbox[t]{3mm}{\multirow{14}{*}{\rotatebox[origin=c]{90}{\small Constrained}}} & \multirow{4}{*}{1} & CUNI-MT & transfer2 & $31.6$ & $0.600$ \\
				& & CUNI-MT & base & $31.4$ & $0.596$ \\
				& & CUNI-MT & transfer1 & $31.3$ & $0.595$ \\
				& & PROMT & multilingual & $31.1$ & $0.599$ \\
				\cmidrule(lr){2-6}
				& 2 & ETRANSLATION & basetr & $30.4$ & $0.593$ \\
				\cmidrule(lr){2-6}
				& \multirow{2}{*}{3} & CUNI-MT & transfer3 & $29.8$ & $0.584$ \\
				& & LC & multilingual & $29.5$ & $0.584$ \\
				\cmidrule(lr){2-6}
				& \multirow{2}{*}{4} & Baseline & Transformer & $28.1$ & $0.573$ \\
				& & LC & transformer & $26.7$ & $0.556$ \\
				\cmidrule(lr){2-6}
				& 5 & TARJAMA-AI & base3 & $25.6$ & $0.564$ \\
				\cmidrule(lr){2-6}
				& 6 & TARJAMA-AI & base2 & $25.0$ & $0.559$ \\
				\cmidrule(lr){2-6}
				& 7 & CUNI-MTIR & r1 & $19.7$ & $0.494$ \\
				\cmidrule(lr){2-6}
				& \multirow{2}{*}{8} & Baseline & RNN & $17.9$ & $0.479$ \\
				& & TARJAMA-AI & base & $17.7$ & $0.488$ \\
				\cmidrule(lr){2-6}\morecmidrules\cmidrule(lr){2-6}
				\parbox[t]{3mm}{\multirow{4}{*}{\rotatebox[origin=c]{90}{\small Unconstrained}}} & 1 & ETRANSLATION & wmtfinetune & $44.4$ & $0.686$ \\
				\cmidrule(lr){2-6}
				& 2 & ETRANSLATION & wmt & $44.1$ & $0.683$ \\
				\cmidrule(lr){2-6}
				& 3 & PROMT & transformer & $41.2$ & $0.666$ \\
				\cmidrule(lr){2-6}
				& 4 & CUNI-MTIR & r1 & $20.0$ & $0.499$ \\
				\bottomrule
			\end{tabular}
	\end{minipage}}
	\hspace{-60pt}
	\resizebox{0.7\textwidth}{!}{\begin{minipage}{\textwidth}
			\begin{tabular}{c c c c c c}
				\multicolumn{6}{c}{\textbf{English{\textendash}French}} \\
				\toprule
				& \textbf{Rank} & \textbf{Team} & \textbf{Description} & \textbf{BLEU [$\uparrow$]} & \textbf{chrF [$\uparrow$]} \\
				\cmidrule(lr){2-2}\cmidrule(lr){3-3}\cmidrule(lr){4-4}\cmidrule(lr){5-5}\cmidrule(lr){6-6}
				\parbox[t]{3mm}{\multirow{12}{*}{\rotatebox[origin=c]{90}{\small Constrained}}} & 1 & PROMT & multilingual & $49.6$ & $0.711$ \\
				\cmidrule(lr){2-6}
				& \multirow{8}{*}{2} & ETRANSLATION & small & $49.1$ & $0.707$ \\
				& & LC & multilingual & $49.0$ & $0.705$ \\
				& & LC & transformer & $48.9$ & $0.703$ \\
				& & CUNI-MT & base & $48.4$ & $0.703$ \\
				& & CUNI-MT & multiling & $48.0$ & $0.700$ \\
				& & ETRANSLATION & big & $47.4$ & $0.695$ \\
				& & Baseline & Transformer & $47.3$ & $0.693$ \\
				& & CUNI-MT & transfer2 & $47.1$ & $0.693$ \\
				\cmidrule(lr){2-6}
				& 3 & LIMSI & trans & $43.5$ & $0.660$ \\
				\cmidrule(lr){2-6}
				& 4 & CUNI-MTIR & r1 & $34.9$ & $0.605$ \\
				\cmidrule(lr){2-6}
				& - & Baseline & RNN & $34.3$ & $0.596$ \\
				\cmidrule(lr){2-6}
				& 5 & TARJAMA-AI & base & $26.8$ & $0.567$ \\
				\cmidrule(lr){2-6}
				& 6 & ACCENTURE & mbart & $15.8$ & $0.464$ \\
				\cmidrule(lr){2-6}\morecmidrules\cmidrule(lr){2-6}
				\parbox[t]{3mm}{\multirow{9}{*}{\rotatebox[origin=c]{90}{\small Unconstrained}}} & 1 & PROMT & transformer & $59.5$ & $0.767$ \\
				\cmidrule(lr){2-6}
				& 2 & ETRANSLATION & gen & $52.9$ & $0.742$ \\
				\cmidrule(lr){2-6}
				& 3 & LIMSI & indom & $51.2$ & $0.721$ \\
				\cmidrule(lr){2-6}
				& \multirow{4}{*}{4} & ETRANSLATION & phwt	& $50.1$ & $0.724$ \\
				& & LIMSI & trans & $49.3$ & $0.710$ \\
				& & LIMSI & bert & $49.3$ & $0.703$ \\
				& & LIMSI & mlia & $48.5$ & $0.705$ \\
				\cmidrule(lr){2-6}
				& 5 & ETRANSLATION & eufl & $47.9$ & $0.712$ \\
				\cmidrule(lr){2-6}
				& 6 & CUNI-MTIR & r1 & $33.0$ & $0.590$ \\
				\bottomrule
			\end{tabular}
	\end{minipage}}
	
	\begin{minipage}{\textwidth}
		$ $
	\end{minipage}
	
	\hspace{-20pt}
	\resizebox{0.7\textwidth}{!}{\begin{minipage}{\textwidth}
			\begin{tabular}{c c c c c c}
				\multicolumn{6}{c}{\textbf{English{\textendash}Spanish}} \\
				\toprule
				& \textbf{Rank} & \textbf{Team} & \textbf{Description} & \textbf{BLEU [$\uparrow$]} & \textbf{chrF [$\uparrow$]} \\
				\cmidrule(lr){2-2}\cmidrule(lr){3-3}\cmidrule(lr){4-4}\cmidrule(lr){5-5}\cmidrule(lr){6-6}
				\parbox[t]{3mm}{\multirow{11}{*}{\rotatebox[origin=c]{90}{\small Constrained}}} & 1 & PROMT & multilingual & $48.3$ & $0.702$ \\
				\cmidrule(lr){2-6}
				& \multirow{6}{*}{2} & CUNI-MT & transfer1 & $47.9$ & $0.699$ \\
				& & CUNI-MT & transfer2 & $47.6$ & $0.698$ \\
				& & LC & multilingual & $47.5$ & $0.695$ \\
				& & Baseline & Transformer & $47.4$ & $0.694$ \\
				& & CUNI-MT & multiling & $47.3$ & $0.692$ \\
				& & CUNI-MT & base & $47.3$ & $0.691$ \\
				\cmidrule(lr){2-6}
				& - & Baseline & RNN & $35.6$ & $0.609$ \\
				\cmidrule(lr){2-6}
				& 3 & CUNI-MTIR & r1 & $32.9$ & $0.591$ \\
				\cmidrule(lr){2-6}
				& 4 & TARJAMA-AI & base & $30.9$ & $0.593$ \\
				\cmidrule(lr){2-6}
				& 5 & ACCENTURE & mbart & $17.4$ & $0.474$ \\
				\cmidrule(lr){2-6}\morecmidrules\cmidrule(lr){2-6}
				\parbox[t]{3mm}{\multirow{2}{*}{\rotatebox[origin=c]{90}{\small Uncon.}}} & 1 & PROMT & transformer & 58.2 & 0.762 \\
				\cmidrule(lr){2-6}
				& 2 & CUNI-MTIR & r1 & 32.1 & 0.582 \\
				\bottomrule
			\end{tabular}
	\end{minipage}} 
	\hspace{-60pt}
	\resizebox{0.7\textwidth}{!}{\begin{minipage}{\textwidth}
			\begin{tabular}{c c c c c c}
				\multicolumn{6}{c}{\textbf{English{\textendash}Italian}} \\
				\toprule
				& \textbf{Rank} & \textbf{Team} & \textbf{Description} & \textbf{BLEU [$\uparrow$]} & \textbf{chrF [$\uparrow$]} \\
				\cmidrule(lr){2-2}\cmidrule(lr){3-3}\cmidrule(lr){4-4}\cmidrule(lr){5-5}\cmidrule(lr){6-6}
				\parbox[t]{3mm}{\multirow{7}{*}{\rotatebox[origin=c]{90}{\small Constrained}}} & 1 & PROMT & multilingual & $29.6$ & $0.585$ \\
				\cmidrule(lr){2-6}
				& \multirow{3}{*}{2} & LC & multilingual & $28.4$ & $0.572$ \\
				& & CUNI-MT & transfer2 & $28.3$ & $0.574$ \\
				& & CUNI-MT & multiling & $28.3$ & $0.574$ \\
				\cmidrule(lr){2-6}
				& - & Baseline & Transformer & $26.9$ & $0.560$ \\
				\cmidrule(lr){2-6}
				& 3 & TARJAMA-AI & base & $19.2$ & $0.494$ \\
				\cmidrule(lr){2-6}
				& - & Baseline & RNN & $17.0$ & $0.473$ \\
				\cmidrule(lr){2-6}\morecmidrules\cmidrule(lr){2-6}
				{\small Uncon.} & 1 & PROMT & transformer & 38.0 & 0.642 \\
				\bottomrule
			\end{tabular}
	\end{minipage}}
	
	\begin{minipage}{\textwidth}
		$ $
	\end{minipage}
	
	\hspace{-20pt}
	\resizebox{0.7\textwidth}{!}{\begin{minipage}{\textwidth}
			\begin{tabular}{c c c c c c}
				\multicolumn{6}{c}{\textbf{English{\textendash}Modern Greek}} \\
				\toprule
				& \textbf{Rank} & \textbf{Team} & \textbf{Description} & \textbf{BLEU [$\uparrow$]} & \textbf{chrF [$\uparrow$]} \\
				\cmidrule(lr){2-2}\cmidrule(lr){3-3}\cmidrule(lr){4-4}\cmidrule(lr){5-5}\cmidrule(lr){6-6}
				\parbox[t]{3mm}{\multirow{5}{*}{\rotatebox[origin=c]{90}{\small Constrained}}} & 1 & PROMT & multilingual & $27.2$ & $0.523$ \\
				\cmidrule(lr){2-6}
				& 2 & CUNI-MT & transfer1 & $24.7$ & $0.496$ \\
				\cmidrule(lr){2-6}
				& \multirow{2}{*}{3} & CUNI-MT & base & $24.1$ & $0.484$ \\
				& & Baseline & Transformer & $22.6$ & $0.471$ \\
				\cmidrule(lr){2-6}
				& - & Baseline & RNN & $12.8$ & $0.365$ \\
				\cmidrule(lr){2-6}\morecmidrules\cmidrule(lr){2-6}
				{\small Uncon.} & 1 & PROMT & transformer & 42.4 & 0.652 \\
				\bottomrule
			\end{tabular}
	\end{minipage}}
	\hspace{-60pt}
	\resizebox{0.7\textwidth}{!}{\begin{minipage}{\textwidth}
			\begin{tabular}{c c c c c c}
				\multicolumn{6}{c}{\textbf{English{\textendash}Swedish}} \\
				\toprule
				& \textbf{Rank} & \textbf{Team} & \textbf{Description} & \textbf{BLEU [$\uparrow$]} & \textbf{chrF [$\uparrow$]} \\
				\cmidrule(lr){2-2}\cmidrule(lr){3-3}\cmidrule(lr){4-4}\cmidrule(lr){5-5}\cmidrule(lr){6-6}
				\parbox[t]{3mm}{\multirow{9}{*}{\rotatebox[origin=c]{90}{\small Constrained}}} & \multirow{3}{*}{1} & PROMT & multilingual & $30.7$ & $0.595$ \\
				& & LC & multilingual & $30.4$ & $0.589$ \\
				& & CUNI-MT & transfer2 & $30.1$ & $0.590$ \\
				\cmidrule(lr){2-6}
				& 2 & CUNI-MT & transfer & $28.5$ & $0.578$ \\
				\cmidrule(lr){2-6}
				& - & Baseline & Transformer & $27.8$ & $0.566$ \\
				\cmidrule(lr){2-6}
				& 3 & CUNI-MT & base & $26.6$ & $0.561$ \\
				\cmidrule(lr){2-6}
				& 4 & CUNI-MTIR & r1 & $25.1$ & $0.541$ \\
				\cmidrule(lr){2-6}
				& - & Baseline & RNN & $19.2$ & $0.481$ \\
				\cmidrule(lr){2-6}
				& 5 & TARJAMA-AI & base & $11.2$ & $0.443$ \\
				\cmidrule(lr){2-6}\morecmidrules\cmidrule(lr){2-6}
				\parbox[t]{3mm}{\multirow{2}{*}{\rotatebox[origin=c]{90}{\small Uncon.}}} &  1 & PROMT & transformer & 41.3 & 0.671 \\
				\cmidrule(lr){2-6}
				& 2 & CUNI-MTIR & r1 & 24.0 & 0.514 \\
				\bottomrule
			\end{tabular}
	\end{minipage}}
\end{table*}

\subsection{Round 1}
\cref{ta:res1} presents the results of the first round. Overall, \emph{multilingual} and \emph{transfer learning} approaches yielded the best results for all language pairs in the constrained scenario. In fact, except for English{\textendash}German (in which they shared the same ranking), \emph{PROMT}'s multilingual approach{\textemdash}which was the only multilingual system trained for all language pairs{\textemdash}achieved the best results in all cases. This approach also used a smaller vocabulary and \emph{SentencePiece} instead of \ac{bpe}.

In general, the differences from one position to the next one were of a few points (according to both metrics), with a case (English{\textendash}French) in which there are two points of difference (according to BLEU) between the first and last approaches of the same ranking. Our baselines worked well as delimiters: more sophisticated approaches generally ranked above our Transformer baselines, while the rest ranked either between them or below the RNN baselines. Moreover, the RNN baselines established the limit before a significant drop in translation quality between approaches of one position in the ranking and the next position (sometimes it is the exact limit, while other times there is a cluster above it of a similar quality).

Regarding the unconstrained scenario, it had less participation than the constrained one. With an exemption (\emph{ETRANSLATION}'s approaches based on their WMT system \citep{Oravecz20} yielded the best results for English{\textendash}German), \emph{PROMT}'s multilingual approach achieved the best results for all language pairs. In general, approaches were similar to the constrained ones but using additional external data. Additionally, due to the use of external data, the best unconstrained systems yielded around 10 BLEU points and 7 ChrF points of improvement compared to the best constrained systems for each language pair.

\subsubsection*{English{\textendash}German}
For this language pair, 12 different systems from 6 participants were submitted to the constrained scenario, and 4 different systems from 3 participants were submitted to the unconstrained one.

The best results for the constrained scenario were based on transfer learning, standard \ac{nmt} with back-translation and multilingual \ac{nmt}. Regarding the unconstrained systems, the best results were yielded by \emph{ETRANSLATION}'s WMT systems, one of which had been fine-tuned with the in-domain data. It is worth noting how, despite being trained using exclusively out-of-domain data, its WMT approach achieved around 13 BLEU points of improvement over the best constrained system. We discussed this phenomenon during the campaign wrap-up workshop and came to the conclusion that, despite working in a very specific domain (Covid-19 related documents), the sub-genre of this domain\footnote{Note that the target audience of Covid-19 MLIA is the general public. Thus, while documents are indeed Covid-19 related, they lean towards information aimed at the citizens instead of  scientific or medical experts.} is more closely related to the news domain of WMT than expected.

\subsubsection*{English{\textendash}French}
For this language pair, 12 different systems from 8 participants were submitted to the constrained scenario, and 8 different systems from 4 participants to the unconstrained one. Best results for both scenarios were yielded by multilingual approaches.

\subsubsection*{English{\textendash}Spanish}
For this language pair, 9 different systems from 6 participants were submitted to the constrained scenario, and only 2 different systems from 2 participants for the unconstrained one.

For the constrained scenario, the best approaches were based on multilingual systems and transfer learning. Regarding the unconstrained scenario, the two submissions consisted on a multilingual system and a Transformer-based approach.

\subsubsection*{English{\textendash}Italian}
For this language pair, 5 different systems from 4 participants were submitted for the constrained scenario, and a single system to the unconstrained one. In all scenarios, best approaches were based on multilingual systems.

\subsubsection*{English{\textendash}Modern Greek}
For this language pair, 3 different systems from 2 participants were submitted for the constrained scenario, and a single system was submitted to the unconstrained one. Thus, this was the language pair with less participation. According to participant's reports, this was mostly due to Modern Greek using a different alphabet. 

Once more, best constrained approaches were based on multilingual systems and transfer learn. In the case of the unconstrained scenario, a single multilingual system was submitted.

\subsubsection*{English{\textendash}Swedish}
For this language pair, 7 different systems from 5 participants were submitted to the constrained scenario, and two system from two participants to the unconstrained one.

For the constrained scenario, best results were obtained using multilingual systems and transfer learning. The submissions of the unconstrained scenario were based on a multilingual system and a Transformer-based approach.

\begin{table*}[!]
	\caption{Results of the second round, divided by categories. Systems are ranked according to BLEU. Lines indicate clusters according to ART. Systems within a cluster are considered tied and, thus, are ranked equally. The baseline \emph{Transformer+} corresponds to the one trained using also the data from round 1.}
	\label{ta:res2}
	\vspace{-5pt}
	\hspace{-25pt}
	\resizebox{0.5\textwidth}{!}{\begin{minipage}{\textwidth}
			\begin{tabular}{c c c c c c c}
				\multicolumn{7}{c}{\textbf{English{\textendash}German}} \\
				\toprule
				& \textbf{Rank} & \textbf{Team} & \textbf{Description} & \textbf{BLEU [$\uparrow$]} & \textbf{TER [$\downarrow$]} & \textbf{BEER [$\uparrow$]} \\
				\cmidrule(lr){2-2}\cmidrule(lr){3-3}\cmidrule(lr){4-4}\cmidrule(lr){5-5}\cmidrule(lr){6-6}\cmidrule(lr){7-7}
				\parbox[t]{3mm}{\multirow{14}{*}{\rotatebox[origin=c]{90}{\small Constrained}}} & \multirow{5}{*}{1} & LC & 5lang-ft-avg & $40.3$ & $48.4$ & $66.8$ \\
				& & ETRANSLATION & ensembleFT & $39.9$ & $48.2$ & $66.8$ \\
				& & LC & 5lang-ft & $39.8$ & $48.9$ & $66.5$ \\
				& & ETRANSLATION & ensemble & $39.7$ & $48.4$ & $66.6$ \\
				& & LC & 1lang & $39.7$ & $50.1$ & $65.9$ \\
				\cmidrule(lr){2-7}
				& 2 & PROMT & multilingual-model-round2-tuned-de & $39.6$ & $47.7$ & $66.8$ \\
				\cmidrule(lr){2-7}
				& 3 & LC & 7lang & $38.6$ & $50.0$ & $65.8$ \\
				\cmidrule(lr){2-7}
				& 4 & PROMT & multilingual-model-round2 & $39.6$ & $49.6$ & $65.7$ \\
				\cmidrule(lr){2-7}
				& \multirow{2}{*}{-} & Baseline & Transformer & $34.9$ & $51.7$ & $63.9$ \\
				& & Baseline & Transformer+ & $34.8$ & $51.8$ & $63.7$ \\
				\cmidrule(lr){2-7}
				& 5 & CUNI-MT & transfer & $31.8$ & $54.6$ & $61.8$ \\
				\cmidrule(lr){2-7}
				& 6 & PROMT & multilingual-model-round1	& $28.7$ & $57.7$ & $60.8$ \\
				\cmidrule(lr){2-7}
				& 7 & CUNI-MT & transfer2 & $27.5$ & $60.4$ & $59.8$ \\
				\cmidrule(lr){2-7}
				& 8 & CUNI-MT & multiling & $27.0$ & $60.9$ & $59.2$ \\
				\cmidrule(lr){2-7}\morecmidrules\cmidrule(lr){2-7}
				\parbox[t]{3mm}{\multirow{5}{*}{\rotatebox[origin=c]{90}{\small Unconstrained}}} & 1 & ETRANSLATION & wmtFT & $45.7$ & $43.0$ & $70.4$ \\
				\cmidrule(lr){2-7}
				& 2 & PROMT & Transformer & $40.4$ & $46.9$ & $67.9$ \\
				\cmidrule(lr){2-7}
				& 3 & ETRANSLATION & singlebigTr & $40.0$ & $48.4$ & $66.9$ \\
				\cmidrule(lr){2-7}
				& 4 & ETRANSLATION & eTstandardengine & $35.4$ & $52.7$ & $64.6$ \\
				\bottomrule
			\end{tabular}
	\end{minipage}}
	\hspace{25pt}
	\resizebox{0.5\textwidth}{!}{\begin{minipage}{\textwidth}
			\begin{tabular}{c c c c c c c}
				\multicolumn{7}{c}{\textbf{English{\textendash}French}} \\
				\toprule
				& \textbf{Rank} & \textbf{Team} & \textbf{Description} & \textbf{BLEU [$\uparrow$]} & \textbf{TER [$\downarrow$]} & \textbf{BEER [$\uparrow$]} \\
				\cmidrule(lr){2-2}\cmidrule(lr){3-3}\cmidrule(lr){4-4}\cmidrule(lr){5-5}\cmidrule(lr){6-6}\cmidrule(lr){7-7}
				\parbox[t]{3mm}{\multirow{11}{*}{\rotatebox[origin=c]{90}{\small Constrained}}} & \multirow{4}{*}{1} & ETRANSLATION & 2 & $58.3$ & $33.8$ & $75.1$ \\
				& & ETRANSLATION & 1 & $57.9$ & $34.0$ & $75.0$ \\
				& & LC & 1lang & $57.2$ & $34.9$ & $74.5$ \\
				& & PROMT & multilingual-model-round2-tuned-fr & $57.1$ & $34.1$ & $74.8$ \\
				\cmidrule(lr){2-7}
				& 2 & CdT-ASL & only-round2-data & $56.9$ & $34.6$ & $74.5$ \\
				\cmidrule(lr){2-7}
				& \multirow{2}{*}{3} & LC & 7lang & $55.8$ & $35.7$ & $73.9$ \\
				&  & PROMT & multilingual-model-round2 & $55.4$ & $35.2$ & $74.0$ \\
				\cmidrule(lr){2-7}
				& - & Baseline & Transformer & $54.4$ & $35.9$ & $73.4$ \\
				\cmidrule(lr){2-7}
				& - & Baseline & Transformer+ & $53.7$ & $36.7$ & $73.0$ \\
				\cmidrule(lr){2-7}
				& 4 & PROMT & multilingual-model-round1	& $45.4$ & $43.1$ & $68.9$ \\
				\cmidrule(lr){2-7}
				& 5 & CUNI-MT & multiling & $44.1$ & $45.6$ & $67.6$ \\
				\cmidrule(lr){2-7}\morecmidrules\cmidrule(lr){2-7}
				\parbox[t]{3mm}{\multirow{5}{*}{\rotatebox[origin=c]{90}{\small Unconstrained}}} & \multirow{2}{*}{1} & PROMT & Transformer & $57.1$ & $34.5$ & $74.8$ \\
				&  & ETRANSLATION & generaldenorm & $56.9$ & $34.8$ & $74.5$ \\
				\cmidrule(lr){2-7}
				& \multirow{2}{*}{2} & ETRANSLATION & general & $49.9$ & $38.8$ & $72.0$ \\
				&  & CdT-ASL & only-cdt-data & $49.7$ & $40.0$ & $71.3$ \\
				\cmidrule(lr){2-7}
				& 3 & ETRANSLATION & formal & $43.5$ & $44.6$ & $68.0$ \\
				\bottomrule
				\multicolumn{7}{c}{ } \\
				\multicolumn{7}{c}{ } \\
				\multicolumn{7}{c}{ } \\
			\end{tabular}
	\end{minipage}}
	
	\begin{minipage}{\textwidth}
		$ $
	\end{minipage}
	
	\hspace{-25pt}
	\resizebox{0.5\textwidth}{!}{\begin{minipage}{\textwidth}
			\begin{tabular}{c c c c c c c}
				\multicolumn{7}{c}{\textbf{English{\textendash}Spanish}} \\
				\toprule
				& \textbf{Rank} & \textbf{Team} & \textbf{Description} & \textbf{BLEU [$\uparrow$]} & \textbf{TER [$\downarrow$]} & \textbf{BEER [$\uparrow$]} \\
				\cmidrule(lr){2-2}\cmidrule(lr){3-3}\cmidrule(lr){4-4}\cmidrule(lr){5-5}\cmidrule(lr){6-6}\cmidrule(lr){7-7}
				\parbox[t]{3mm}{\multirow{13}{*}{\rotatebox[origin=c]{90}{\small Constrained}}} & \multirow{4}{*}{1} & LC & 1lang-avg & $56.6$ & $33.7$ & $75.2$ \\
				& & ETRANSLATION & 2 & $56.1$ & $33.5$ & $75.2$ \\
				& & ETRANSLATION & 1 & $56.1$ & $33.5$ & $75.2$ \\
				& & LC & 5lang-ft-avg & $56.0$ & $33.8$ & $75.1$ \\
				\cmidrule(lr){2-7}
				& \multirow{3}{*}{2} & CdT-ASL & only-round2-data & $55.4$ & $34.1$ & $74.6$ \\
				& & LC & 7lang & $55.3$ & $34.4$ & $74.8$ \\
				& & PROMT & multilingual-model-round2-tuned-es & $54.9$ & $33.9$ & $74.9$ \\
				\cmidrule(lr){2-7}
				& 3 & PROMT & multilingual-model-round2 & $53.8$ & $34.5$ & $74.3$ \\
				\cmidrule(lr){2-7}
				& - & Baseline & Transformer & $53.3$ & $35.2$ & $74.0$ \\
				\cmidrule(lr){2-7}
				& - & Baseline & Transformer+ & $51.8$ & $36.1$ & $73.3$ \\
				\cmidrule(lr){2-7}
				& 4 & CUNI-MT & transfer & $48.4$ & $39.3$ & $71.2$ \\
				\cmidrule(lr){2-7}
				& 5 & PROMT & multilingual-model-round1 & $45.1$ & $41.2$ & $69.9$ \\
				\cmidrule(lr){2-7}
				& 6 & CUNI-MT & multiling & $42.1$ & $45.9$ & $67.4$ \\
				\cmidrule(lr){2-7}\morecmidrules\cmidrule(lr){2-7}
				\parbox[t]{3mm}{\multirow{4}{*}{\rotatebox[origin=c]{90}{\small Unconstrain.}}} & \multirow{2}{*}{1} & ETRANSLATION & 2 & $56.5$ & $33.2$ & $75.4$ \\
				&  & ETRANSLATION & 1 & $56.0$ & $33.5$ & $75.2$ \\
				\cmidrule(lr){2-7}
				& 2 & PROMT & Transformer & $53.2$ & $35.0$ & $74.6$ \\
				\cmidrule(lr){2-7}
				& 3 & CdT-ASL & only-cdt-data & $51.4$ & $37.0$ & $72.9$ \\
				\bottomrule
				\multicolumn{7}{c}{ } \\
				\multicolumn{7}{c}{ } \\
			\end{tabular}
	\end{minipage}}
	\hspace{25pt}
	\resizebox{0.5\textwidth}{!}{\begin{minipage}{\textwidth}
			\begin{tabular}{c c c c c c c}
				\multicolumn{7}{c}{\textbf{English{\textendash}Italian}} \\
				\toprule
				& \textbf{Rank} & \textbf{Team} & \textbf{Description} & \textbf{BLEU [$\uparrow$]} & \textbf{TER [$\downarrow$]} & \textbf{BEER [$\uparrow$]} \\
				\cmidrule(lr){2-2}\cmidrule(lr){3-3}\cmidrule(lr){4-4}\cmidrule(lr){5-5}\cmidrule(lr){6-6}\cmidrule(lr){7-7}
				\parbox[t]{3mm}{\multirow{13}{*}{\rotatebox[origin=c]{90}{\small Constrained}}} & \multirow{2}{*}{1} & LC & 5lang-ov-ft-avg	 & $48.9$ & $40.3$ & $70.2$ \\
				& & PROMT & multilingual-model-round2-tuned-it & $48.3$ & $39.5$ & $70.4$ \\
				\cmidrule(lr){2-7}
				& 2 & LC & 5lang-ov & $48.0$ & $40.9$ & $69.8$ \\
				\cmidrule(lr){2-7}
				& \multirow{2}{*}{3} & ETRANSLATION & 4bigTens & $47.0$ & $41.7$ & $69.7$ \\
				& & PROMT & multilingual-model-round2 & $46.8$ & $40.6$ & $69.1$ \\
				\cmidrule(lr){2-7}
				& 4 & ETRANSLATION & 4bigTensFT & $46.7$ & $42.2$ & $68.9$ \\
				\cmidrule(lr){2-7}
				& 5 & LC & 1lang & $45.3$ & $44.1$ & $67.8$ \\
				\cmidrule(lr){2-7}
				& - & Baseline & Transformer+ & $43.5$ & $44.0$ & $67.9$ \\
				\cmidrule(lr){2-7}
				& - & Baseline & Transformer & $42.9$ & $44.3$ & $67.1$ \\
				\cmidrule(lr){2-7}
				& 6 & CUNI-MT & transfer & $38.6$ & $48.1$ & $64.6$ \\
				\cmidrule(lr){2-7}
				& 7 & CdT-ASL & only-round2-data & $37.9$ & $51.9$ & $62.6$ \\
				\cmidrule(lr){2-7}
				& 8 & PROMT & multilingual-model-round1 & $37.6$ & $48.8$ & $65.0$ \\
				\cmidrule(lr){2-7}
				& 9 & CUNI-MT & multiling & $35.2$ & $53.1$ & $62.5$ \\
				\cmidrule(lr){2-7}\morecmidrules\cmidrule(lr){2-7}
				\parbox[t]{3mm}{\multirow{5}{*}{\rotatebox[origin=c]{90}{\small Unconstrained}}} & 1 & ETRANSLATION & 4bigTens & $50.1$ & $39.0$ & $71.0$ \\
				\cmidrule(lr){2-7}
				& 2 & ETRANSLATION & 4bigTensnorm & $49.9$ & $39.4$ & $70.9$ \\
				\cmidrule(lr){2-7}
				& \multirow{2}{*}{3} & CdT-ASL & round2-data & $49.0$ & $39.9$ & $70.5$ \\
				&  & PROMT & Transformer & $47.8$ & $40.0$ & $70.6$ \\
				\cmidrule(lr){2-7}
				& 4 & CdT-ASL & only-cdt-data & $45.2$ & $43.3$ & $68.8$ \\
				\bottomrule
			\end{tabular}
	\end{minipage}}
	
	\begin{minipage}{\textwidth}
		$ $
	\end{minipage}
	
	\hspace{-25pt}
	\resizebox{0.5\textwidth}{!}{\begin{minipage}{\textwidth}
			\begin{tabular}{c c c c c c c}
				\multicolumn{7}{c}{\textbf{English{\textendash}Modern Greek}} \\
				\toprule
				& \textbf{Rank} & \textbf{Team} & \textbf{Description} & \textbf{BLEU [$\uparrow$]} & \textbf{TER [$\downarrow$]} & \textbf{BEER [$\uparrow$]} \\
				\cmidrule(lr){2-2}\cmidrule(lr){3-3}\cmidrule(lr){4-4}\cmidrule(lr){5-5}\cmidrule(lr){6-6}\cmidrule(lr){7-7}
				\parbox[t]{3mm}{\multirow{14}{*}{\rotatebox[origin=c]{90}{\small Constrained}}} & 1 & PROMT & multilingual-model-round2-tuned-el & $45.1$ & $42.3$ & $67.8$ \\
				\cmidrule(lr){2-7}
				& 2 & LC & 7lang-ov-ft-avg & $44.7$ & $43.8$ & $67.2$ \\
				\cmidrule(lr){2-7}
				& 3 & LC & 7lang-ov & $44.2$ & $44.1$ & $67.0$ \\
				\cmidrule(lr){2-7}
				& \multirow{2}{*}{4} & LC & 7lang & $43.2$ & $44.8$ & $66.5$ \\
				& & PROMT & multilingual-model-round2 & $42.1$ & $44.3$ & $66.3$ \\
				\cmidrule(lr){2-7}
				& 5 & ETRANSLATION & 1 & $41.7$ & $46.2$ & $65.5$ \\
				\cmidrule(lr){2-7}
				& 6 & LC & 1lang & $41.2$ & $47.3$ & $64.8$ \\
				\cmidrule(lr){2-7}
				& - & Baseline & Transformer+ & $39.8$ & $46.9$ & $64.7$ \\
				\cmidrule(lr){2-7}
				& - & Baseline & Transformer & $38.5$ & $48.2$ & $63.7$ \\
				\cmidrule(lr){2-7}
				& \multirow{2}{*}{7} & ETRANSLATION & 2 & $34.9$ & $53.2$ & $60.8$ \\
				&  & CUNI-MT & transfer & $34.9$ & $51.6$ & $61.3$ \\
				\cmidrule(lr){2-7}
				& \multirow{3}{*}{8} & CdT-ASL & only-round2-data & $32.9$ & $56.6$ & $59.0$ \\
				&  & CUNI-MT & multiling & $32.4$ & $56.1$ & $59.5$ \\
				&  & PROMT & multilingual-model-round1	 & $31.4$ & $55.2$ & $59.5$ \\
				\cmidrule(lr){2-7}\morecmidrules\cmidrule(lr){2-7}
				\parbox[t]{3mm}{\multirow{4}{*}{\rotatebox[origin=c]{90}{\small Unconst.}}} & 1 & PROMT & Transformer & $44.4$ & $44.0$ & $67.2$ \\
				\cmidrule(lr){2-7}
				& 2 & ETRANSLATION & 2 & $44.3$ & $43.9$ & $66.9$ \\
				\cmidrule(lr){2-7}
				& 3 & ETRANSLATION & 1 & $43.1$ & $44.7$ & $66.3$ \\
				\cmidrule(lr){2-7}
				& 4 & CdT-ASL & only-cdt-data & $37.5$ & $50.0$ & $63.7$ \\
				\bottomrule
			\end{tabular}
	\end{minipage}}
	\hspace{25pt}
	\resizebox{0.5\textwidth}{!}{\begin{minipage}{\textwidth}
			\begin{tabular}{c c c c c c c}
				\multicolumn{7}{c}{\textbf{English{\textendash}Swedish}} \\
				\toprule
				& \textbf{Rank} & \textbf{Team} & \textbf{Description} & \textbf{BLEU [$\uparrow$]} & \textbf{TER [$\downarrow$]} & \textbf{BEER [$\uparrow$]} \\
				\cmidrule(lr){2-2}\cmidrule(lr){3-3}\cmidrule(lr){4-4}\cmidrule(lr){5-5}\cmidrule(lr){6-6}\cmidrule(lr){7-7}
				\parbox[t]{3mm}{\multirow{14}{*}{\rotatebox[origin=c]{90}{\small Constrained}}} & \multirow{4}{*}{1} & ETRANSLATION & 4bigTens & $22.7$ & $72.7$ & $48.2$ \\
				&  & LC & 5lang-ov-ft-avg & $22.0$ & $71.7$ & $49.2$ \\
				&  & PROMT & multilingual-model-round2-tuned-sv & $21.8$ & $69.3$ & $49.7$ \\
				&  & LC & 5lang-ov-r2-data	 & $21.8$ & $71.5$ & $49.4$ \\
				\cmidrule(lr){2-7}
				& 2 & PROMT & multilingual-model-round2 & $20.4$ & $70.7$ & $48.9$ \\
				\cmidrule(lr){2-7}
				& 3 & CdT-ASL & only-round2-data & $20.3$ & $75.3$ & $46.5$ \\
				\cmidrule(lr){2-7}
				& - & Baseline & Transformer+ & $19.5$ & $72.2$ & $48.1$ \\
				\cmidrule(lr){2-7}
				& 4 & LC & 7lang-ov-r1-data & $18.3$ & $74.9$ & $47.4$ \\
				\cmidrule(lr){2-7}
				& 5 & LC & 5lang-r1-data & $17.7$ & $75.6$ & $47.0$ \\
				\cmidrule(lr){2-7}
				& 6 & PROMT & multilingual-model-round1 & $17.2$ & $75.3$ & $46.7$ \\
				\cmidrule(lr){2-7}
				& \multirow{2}{*}{7} & LC & 1lang-r1-data & $16.7$ & $78.9$ & $45.4$ \\
				&  & Baseline & Transformer & $15.3$ & $77.5$ & $44.4$ \\
				\cmidrule(lr){2-7}
				& \multirow{2}{*}{8} & CUNI-MT & multiling & $14.7$ & $79.3$ & $45.1$ \\
				&  & CUNI-MT & transfer & $13.9$ & $76.5$ & $43.5$ \\
				\cmidrule(lr){2-7}\morecmidrules\cmidrule(lr){2-7}
				\parbox[t]{3mm}{\multirow{3}{*}{\rotatebox[origin=c]{90}{\small Unconst.}}} & 1 & ETRANSLATION & 4bigTens & $23.3$ & $70.2$ & $50.0$ \\
				\cmidrule(lr){2-7}
				& \multirow{2}{*}{2} & CdT-ASL & only-cdt-data & $21.3$ & $72.7$ & $48.7$ \\
				&  & PROMT & Transformer & $21.0$ & $71.3$ & $49.3$ \\
				\bottomrule
				\multicolumn{7}{c}{ } \\
				\multicolumn{7}{c}{ } \\
			\end{tabular}
	\end{minipage}}
	
	\begin{minipage}{\textwidth}
		$ $
	\end{minipage}
	
	\hspace{90pt}
	\resizebox{0.5\textwidth}{!}{\begin{minipage}{\textwidth}
			\begin{tabular}{c c c c c c c}
				\multicolumn{7}{c}{ } \\
				\multicolumn{7}{c}{\textbf{English{\textendash}Arabic}} \\
				\toprule
				& \textbf{Rank} & \textbf{Team} & \textbf{Description} & \textbf{BLEU [$\uparrow$]} & \textbf{TER [$\downarrow$]} & \textbf{BEER [$\uparrow$]} \\
				\cmidrule(lr){2-2}\cmidrule(lr){3-3}\cmidrule(lr){4-4}\cmidrule(lr){5-5}\cmidrule(lr){6-6}\cmidrule(lr){7-7}
				\parbox[t]{3mm}{\multirow{9}{*}{\rotatebox[origin=c]{90}{\small Constrained}}} & 1 & LC & 7lang-ov & $25.1$ & $64.7$ & $57.6$ \\
				\cmidrule(lr){2-7}
				& 2 & PROMT & multilingual-model-round2-tuned-ar & $22.9$ & $62.9$ & $56.5$ \\
				\cmidrule(lr){2-7}
				& \multirow{2}{*}{3} & LC & 7lang & $22.0$ & $67.4$ & $55.8$ \\
				&  & PROMT & multilingual-model-round2 & $21.7$ & $63.8$ & $55.9$ \\
				\cmidrule(lr){2-7}
				& \multirow{3}{*}{4} & CUNI-MT & transfer & $19.1$ & $68.7$ & $52.9$ \\
				&  & LC & 1lang & $19.1$ & $73.8$ & $53.0$ \\
				&  & Baseline & Transformer & $18.8$ & $69.3$ & $52.3$ \\
				\cmidrule(lr){2-7}
				& 5 & CUNI-MT & multiling & $17.0$ & $75.2$ & $51.3$ \\
				\cmidrule(lr){2-7}
				& 6 & CdT-ASL & only-round2-data & $15.9$ & $77.9$ & $48.7$ \\
				\cmidrule(lr){2-7}\morecmidrules\cmidrule(lr){2-7}
				{\small Un.} & 1 & PROMT & Transformer & $31.4$ & $54.2$ & $62.3$ \\
				\bottomrule
			\end{tabular}
	\end{minipage}}
\end{table*}

\subsection{Round 2}
\cref{ta:res2} presents the results of the second round. With the exception of English{\textendash}French, in which \emph{monolingual approaches} achieved the best results, \emph{multilingual approaches} yielded the best performances. In the case of English{\textemdash}German, \emph{system ensembling} also ranked at first position.

In general, the differences from one position to the next one were of a few points (according to all metrics). Our baselines worked well as delimiters: more sophisticated approaches generally ranked above our baselines, while the following cluster after them obtained the highest quality drop between consecutive ranks.

\subsubsection*{English{\textendash}German}
For this language pair, 12 different systems from 4 participants were submitted to the constrained scenario, and 4 different systems from 2 participants were submitted to the unconstrained one. Best results were achieved by approaches based on multilingual systems and model ensembling.

\subsubsection*{English{\textendash}French}
For this language pair, 9 different systems from 5 participants were submitted to the constrained scenario, and 5 different systems from 3 participants to the unconstrained one. Best results for both scenarios were yielded by monolingual approaches, with the exception of PROMT's multilingual model (which ranked at the first cluster for the constrained scenario).

\subsubsection*{English{\textendash}Spanish}
For this language pair, 11 different systems from 5 participants were submitted to the constrained scenario, and 4 different systems from 3 participants for the unconstrained one. Best approaches were based both on monolingual and multilingual models.

\subsubsection*{English{\textendash}Italian}
For this language pair, 11 different systems from 5 participants were submitted for the constrained scenario, and 5 different system from 3 participants to the unconstrained one. In all scenarios, best approaches were based on multilingual models.

\subsubsection*{English{\textendash}Modern Greek}
For this language pair, 12 different systems from 5 participants were submitted for the constrained scenario, and 4 different systems from 3 participants to the unconstrained one. Best approaches were based on multilingual models.

\subsubsection*{English{\textendash}Swedish}
For this language pair, 12 different systems from 5 participants were submitted to the constrained scenario, and 3 system from 3 participants to the unconstrained one. Best approaches were based on multilingual models.

\subsubsection*{English{\textendash}Arabic}
For this language pair, 8 different systems from 4 participants were submitted to the constrained scenario, and a single system was submitted to the unconstrained one. This language pair was introduced at this round. Best approaches were based on multilingual models.

\section{Human quality assessment}
\label{se:qa}
Taking into account that the corpora were obtained from crawling (see \cref{se:dg}), it is important to assess the quality of the reference sets. To do so, we selected a subset of the Spanish round 1 corpora and post-edited it with the help of a team of professional translators. This subset consisted of the worst 500 segments according to the alignment probability between source and reference. Overall, translators thought that \emph{the translations in general are good but some are very free adding things that are not in the source or they are too literal}.

As a first step towards assessing the quality of the reference sets, we compared the reference and its post-edited version using human TER (hTER). This metric computes the number of errors between a translation hypothesis and its post-edited version (in this case, between the automatic reference and its post-edited version). Thus, the smallest the value the highest the quality. We obtained a fairly low hTER value ($18.8$), which indicates that the translation quality of the reference is generally good and, thus, is coherent with the translators' opinion.

As a second step, we re-evaluated participant's translations (the corresponding subset only) using both the reference and its post-edited version. \cref{ta:esqa} present the results of the evaluation. In all cases, both metrics show fairly similar results{\textemdash}with a preference towards the reference, which is to be expected since its style is more similar to the training data. Thus, we can conclude that the quality of the reference sets is good enough to be used in an automatic evaluation and that the results obtained in the previous section (\cref{se:res}) are significant.

\begin{table}[!h]
	\caption{Results of evaluating a subset of the Spanish test using either the reference or its post-edited version.}
	\label{ta:esqa}
	\centering
	\resizebox{0.61\textwidth}{!}{\begin{minipage}{\textwidth}
			\begin{tabular}{c c c c c c}
				\toprule
				\multirow{2}{*}{\textbf{Team}} & \multirow{2}{*}{\textbf{Description}} & \multicolumn{2}{c}{\textbf{Reference}} & \multicolumn{2}{c}{\textbf{Post-edition}} \\
				\cmidrule(lr){3-4}\cmidrule(lr){5-6}
				&  & \textbf{BLEU [$\uparrow$]} & \textbf{chrF [$\uparrow$]} & \textbf{BLEU [$\uparrow$]} & \textbf{chrF [$\uparrow$]} \\
				\midrule
				PROMT & multilingual & $45.1$ & $0.682$ & $43.9$ & $0.672$ \\
				CUNI-MT & transfer1 & $46.2$ & $0.686$ & $43.8$ & $0.672$ \\
				CUNI-MT & transfer2 & $46.0$ & $0.686$ & $43.4$ & $0.671$ \\
				LC & multilingual & $45.8$ & $0.684$ & $43.5$ & $0.669$ \\
				Baseline & transformer & $45.4$ & $0.682$ & $43.9$ & $0.670$ \\
				CUNI-MT & multiling & $44.7$ & $0.677$ & $43.0$ & $0.664$ \\
				CUNI-MT & base & $45.0$ & $0.675$ & $42.4$ & $0.660$ \\
				Baseline & RNN & $34.6$ & $0.603$ & $32.3$ & $0.589$ \\
				CUNI-MTIR & r1 & $31.4$ & $0.583$ & $30.8$ & $0.578$ \\
				TARJAMA-AI & base & $29.2$ & $0.583$ & $26.9$ & $0.569$ \\		
				ACCENTURE & mbart & $16.7$ & $0.466$ & $16.0$ & $0.460$ \\
				\bottomrule
			\end{tabular}
	\end{minipage}}
\end{table}

\section{Conclusions}
\label{se:con}
This work presents a community evaluation effort to improve the generation of \ac{mt} systems as a response to a global problem. The initiative consisted of generating specialized corpus for a new and important topic: Covid-19. This initiative was divided into two rounds.

This first round addressed 6 different language pairs and was divided into two scenarios: one in which participants were limited to using only the provided corpora (constrained) and another one in which the use of external tools and data was allowed (unconstrained). 8 different teams took part in this round. Among their approaches, the most successful ones were based on multilingual \ac{mt} and transfer learning, using the Transformer architecture. 

The second round addressed 7 different language pairs and was also divided into constrained and unconstrained scenarios, and engaged 5 different teams. Overall, a focus on the cleaning process of the data yielded great improvements. Most approaches were based on the Transformer and big Transformer architectures. Once more, multilingual models achieved great results, showing to be specially beneficial for languages with less resources. These results make sense due to the fact that Covid-19 corpora are needed to specialize models in the domain, even if they are from another language.

\section*{Acknowledgements}
The Covid-19 MLIA @ Eval initiative has received support from the European Commission, the European Language Resources Coordination (ELRC), European Language Resources Association (ELRA), the European Research Infrastructure for Language Resources and Technology (CLARIN), the CLEF Initiative and the Joint Research Centre (JRC). We gratefully acknowledge the translation team from Pangeanic for their help with the quality assessment.

\bibliographystyle{apalike}
\bibliography{covid}

\begin{thebibliography}{}

\bibitem[Bahdanau et~al., 2015]{Bahdanau15}
Bahdanau, D., Cho, K., and Bengio, Y. (2015).
\newblock Neural machine translation by jointly learning to align and
  translate.
\newblock \textit{arXiv:1409.0473}.

\bibitem[Barrault et~al., 2020]{Barrault20}
Barrault, L., Biesialska, M., Bojar, O., Costa-juss{\`a}, M.~R., Federmann, C.,
  Graham, Y., Grundkiewicz, R., Haddow, B., Huck, M., Joanis, E., Kocmi, T.,
  Koehn, P., Lo, C.-k., Ljube{\v{s}}i{\'c}, N., Monz, C., Morishita, M.,
  Nagata, M., Nakazawa, T., Pal, S., Post, M., and Zampieri, M. (2020).
\newblock Findings of the 2020 conference on machine translation ({WMT}20).
\newblock In {\em Proceedings of the Fifth Conference on Machine Translation},
  pages 1--55.

\bibitem[Gers et~al., 2000]{Gers00}
Gers, F.~A., Schmidhuber, J., and Cummins, F. (2000).
\newblock Learning to forget: Continual prediction with {LSTM}.
\newblock {\em Neural computation}, 12(10):2451--2471.

\bibitem[Jacquet and Verile, 2020]{Jacquet20}
Jacquet, G. and Verile, M. (2020).
\newblock Covid-19 news monitoring with medical information system (medisys).
\newblock European Commission, Joint Research Centre (JRC). PID:
  \url{http://data.europa.eu/89h/bd2f71e7-0551-4f57-8e82-fcfca8c1a462}.

\bibitem[Junczys-Dowmunt et~al., 2018]{mariannmt}
Junczys-Dowmunt, M., Grundkiewicz, R., Dwojak, T., Hoang, H., Heafield, K.,
  Neckermann, T., Seide, F., Germann, U., Fikri~Aji, A., Bogoychev, N.,
  Martins, A. F.~T., and Birch, A. (2018).
\newblock Marian: Fast neural machine translation in {C++}.
\newblock In {\em Proceedings of ACL 2018, System Demonstrations}, pages
  116--121.

\bibitem[Kingma and Ba, 2014]{Kingma14}
Kingma, D.~P. and Ba, J. (2014).
\newblock Adam: A method for stochastic optimization.
\newblock {\em arXiv preprint arXiv:1412.6980}.

\bibitem[{Klein} et~al., 2017]{Klein17}
{Klein}, G., {Kim}, Y., {Deng}, Y., {Senellart}, J., and {Rush}, A.~M. (2017).
\newblock {OpenNMT: Open-Source Toolkit for Neural Machine Translation}.
\newblock In {\em Proceedings of the Association for Computational Linguistics:
  System Demonstration}, pages 67--72.

\bibitem[Kocmi and Bojar, 2018]{cunimttransfer}
Kocmi, T. and Bojar, O. (2018).
\newblock Trivial transfer learning for low-resource neural machine
  translation.
\newblock In {\em Proceedings of the Third Conference on Machine Translation},
  pages 244--252.

\bibitem[Kudo and Richardson, 2018]{sentencepiece}
Kudo, T. and Richardson, J. (2018).
\newblock Sentencepiece: {A} simple and language independent subword tokenizer
  and detokenizer for neural text processing.
\newblock {\em CoRR}, abs/1808.06226.

\bibitem[Lample et~al., 2017]{onlineBT}
Lample, G., Denoyer, L., and Ranzato, M. (2017).
\newblock Unsupervised machine translation using monolingual corpora only.
\newblock {\em CoRR}, abs/1711.00043.

\bibitem[Linge et~al., 2010]{Linge10}
Linge, J., Steinberger, R., Fuart, F., Bucci, S., Belyaeva, J., Gemo, M.,
  Al-Khudhairy, D., Yangarber, R., and van~der Goot, E. (2010).
\newblock {MediSys}: medical information system. in advanced {ICTs} for
  disaster management and threat detection: Collaborative and distributed
  frameworks.
\newblock {\em IGI Global}, pages 131--142.

\bibitem[Liu et~al., 2020]{mbart}
Liu, Y., Gu, J., Goyal, N., Li, X., Edunov, S., Ghazvininejad, M., Lewis, M.,
  and Zettlemoyer, L. (2020).
\newblock Multilingual denoising pre-training for neural machine translation.
\newblock {\em arXiv preprint arXiv:2001.08210}.

\bibitem[Molchanov, 2019]{molchanov-2019-promt}
Molchanov, A. (2019).
\newblock {PROMT} systems for {WMT} 2019 shared translation task.
\newblock In {\em Proceedings of the Fourth Conference on Machine Translation
  (Volume 2: Shared Task Papers, Day 1)}, pages 302--307.

\bibitem[Oravecz et~al., 2020]{Oravecz20}
Oravecz, C., Bontcheva, K., Tihanyi, L., Kolovratnik, D., Bhaskar, B.,
  Lardilleux, A., Klocek, S., and Eisele, A. (2020).
\newblock etranslation's submissions to the wmt 2020 news translation task.
\newblock In {\em Proceedings of the Fifth Conference on Machine Translation},
  pages 254--261.

\bibitem[Papavassiliou et~al., 2013]{Papavassiliou13}
Papavassiliou, V., Prokopidis, P., and Thurmair, G. (2013).
\newblock A modular open-source focused crawler for mining monolingual and
  bilingual corpora from the web.
\newblock In {\em Proceedings of the Sixth Workshop on Building and Using
  Comparable Corpora}, pages 43--51.

\bibitem[Papineni et~al., 2002]{Papineni02}
Papineni, K., Roukos, S., Ward, T., and Zhu, W.-J. (2002).
\newblock {BLEU}: a method for automatic evaluation of machine translation.
\newblock In {\em Proceedings of the Annual Meeting of the Association for
  Computational Linguistics}, pages 311--318.

\bibitem[Popovi{\'c}, 2015]{Popovi15}
Popovi{\'c}, M. (2015).
\newblock chr{F}: character n-gram {F}-score for automatic {MT} evaluation.
\newblock In {\em Proceedings of the Tenth Workshop on Statistical Machine
  Translation}, pages 392--395.

\bibitem[Post, 2018]{Post18}
Post, M. (2018).
\newblock A call for clarity in reporting bleu scores.
\newblock In {\em Proceedings of the Third Conference on Machine Translation},
  pages 186--191.

\bibitem[Riezler and Maxwell, 2005]{Riezler05}
Riezler, S. and Maxwell, J.~T. (2005).
\newblock On some pitfalls in automatic evaluation and significance testing for
  mt.
\newblock In {\em Proceedings of the workshop on intrinsic and extrinsic
  evaluation measures for machine translation and/or summarization}, pages
  57--64.

\bibitem[Roussis et~al., 2022]{Roussis22}
Roussis, D., Papavassiliou, V., Sofianopoulos, S., Prokopidis, P., and
  Piperidis, S. (2022).
\newblock Constructing parallel corpora from covid-19 news using medisys
  metadata.
\newblock Under review at LREC 2022.

\bibitem[Sennrich et~al., 2016]{Sennrich16a}
Sennrich, R., Haddow, B., and Birch, A. (2016).
\newblock Neural machine translation of rare words with subword units.
\newblock In {\em Proceedings of the Annual Meeting of the Association for
  Computational Linguistics}, pages 1715--1725.

\bibitem[Snover et~al., 2006]{Snover06}
Snover, M., Dorr, B., Schwartz, R., Micciulla, L., and Makhoul, J. (2006).
\newblock A study of translation edit rate with targeted human annotation.
\newblock In {\em Proceedings of the Association for Machine Translation in the
  Americas}, pages 223--231.

\bibitem[Stanojevi{\'c} and Sima{'}an, 2014]{Stanojevic14}
Stanojevi{\'c}, M. and Sima{'}an, K. (2014).
\newblock Fitting sentence level translation evaluation with many dense
  features.
\newblock In {\em Proceedings of the 2014 Conference on Empirical Methods in
  Natural Language Processing ({EMNLP})}, pages 202--206.

\bibitem[Sutskever et~al., 2014]{Sutskever14}
Sutskever, I., Vinyals, O., and Le, Q.~V. (2014).
\newblock Sequence to sequence learning with neural networks.
\newblock In {\em Proceedings of the Advances in Neural Information Processing
  Systems}, volume~27, pages 3104--3112.

\bibitem[Szegedy et~al., 2015]{Szegedy15}
Szegedy, C., Liu, W., Jia, Y., Sermanet, P., Reed, S., Anguelov, D., Erhan, D.,
  Vanhoucke, V., and Rabinovich, A. (2015).
\newblock Going deeper with convolutions.
\newblock In {\em Proceedings of the IEEE Conference on Computer Vision and
  Pattern Recognition}, pages 1--9.

\bibitem[Tiedemann, 2012]{opus}
Tiedemann, J. (2012).
\newblock Parallel data, tools and interfaces in {OPUS}.
\newblock In {\em Proceedings of the Eighth International Conference on
  Language Resources and Evaluation ({LREC}'12)}, pages 2214--2218.

\bibitem[Vaswani et~al., 2017]{Vaswani17}
Vaswani, A., Shazeer, N., Parmar, N., Uszkoreit, J., Jones, L., Gomez, A.~N.,
  Kaiser, {\L}., and Polosukhin, I. (2017).
\newblock Attention is all you need.
\newblock In {\em Advances in Neural Information Processing Systems}, pages
  5998--6008.

\bibitem[Zhu et~al., 2020]{zhu2020incorporating}
Zhu, J., Xia, Y., Wu, L., He, D., Qin, T., Zhou, W., Li, H., and Liu, T.-Y.
  (2020).
\newblock Incorporating bert into neural machine translation.
\newblock {\em arXiv preprint arXiv:2002.06823}.

\end{thebibliography}

\end{document}